\documentclass{article}

\usepackage[english]{babel}

\usepackage[letterpaper,top=2cm,bottom=2cm,left=3cm,right=3cm,marginparwidth=1.75cm]{geometry}

\usepackage{amsmath}
\usepackage{graphicx}
\usepackage{color}
\usepackage[colorlinks=true, allcolors=blue]{hyperref}

\title{Towards Deployable RL - What's Broken with RL Research and a Potential Fix}

\author{Shie Mannor\footnote{Technion and Nvidia Research; this writeup represents my own opinions.} 
 \ and Aviv Tamar\footnote{Technion; this writeup represents my own opinions.}}
\date{January 1, 2023} 

\begin{document}
\maketitle

\newcommand{\at}[1]{\textcolor{magenta}{#1}}
\newcommand{\sm}[1]{\textcolor{blue}{#1}}
\newcommand{\todo}[1]{\textcolor{blue}{#1}}

\begin{abstract}
Reinforcement learning (RL) has demonstrated great potential, but is currently full of overhyping and pipe dreams. We point to some difficulties with current research which we feel are endemic to the direction taken by the community. 
To us, the current direction is not likely to lead to ``deployable'' RL: RL that works in practice and can work in practical situations yet still is economically viable. 
We also propose a potential fix to some of the difficulties of the field.
\end{abstract}

\section{Introduction}

Since Deepmind's breakthrough results of applying deep RL to Atari games (2015), RL has been lauded in the AI community, with promises ranging from `a path to AGI', through `the key to self driving cars', and up to `solve all planning problems'. While clear and worthy progress has been made, the current vibe is that aside from solving games, RL is not living up to our expectations. The causes, we believe, are five popular research practices, which were relevant for 2015, but are currently stagnating the field.
As our mission is not to blame anyone (we have surely made our fair share of mistakes), we will keep examples and references to ourselves. Our point is that currently the hype over RL practice and theory is largely unjustified by the facts and that the discipline needs to mature. 

\begin{enumerate}
    \item \textbf{Overfitting to specific benchmarks.} State-of-the-art methods for Atari are by now very different from state-of-the-art methods for Mujoco (or other benchmarks). Nearly every paper is required to show some distinction (under some metric) for one of the popular benchmarks. However, both benchmarks are made up, driving the field further away from algorithms that will be potentially good for practical problems that are not Atari games/robot locomotion. 
    Consequently, research on tweaks that will work most likely only for particular benchmarks is abundant. Furthermore, it is not clear how progress on a benchmark relates to real-world value. In contrast to, say, pretrained ImageNet features or BERT sentence embeddings, which are widely applicable to real world applications, solving RL benchmarks currently does not yield tangible value.
    \item \textbf{Wrong focus.} Most research focuses on sample complexity for a given benchmark. This is hardly the case in practice -- compute can be cheap relative to engineering effort, acquiring labelled demonstrations can significantly speed up learning, and the development process must consider issues like overall system stability, testability, ease of debugging, interpretability, integration with other components/agents, etc. Current benchmarks ignore the deployable nature of situated (RL-driven) agents completely, focusing on algorithms rather than on a system/engineering view. This is most clearly represented in the prevalent OpenAI Gym API, which abstracts away all `system-design' issues for quickly making progress on sample complexity. While initially a good idea, this hampers progress since in many practical problems, just figuring out what are useful states, actions, and rewards is a critical component of the development process.
    \item \textbf{Detached Theory.} What should be the role of theory in modern RL research? There is no real system with small finite state and action spaces. Ideally, theory should help understand phenomena observed in practice, and suggest algorithmic ideas. While ``nothing is more practical than a good theory" (Lewin, 1952), useful theory seems to be quite rare. Some reasons are: regret minimization is overly pessimistic; there is a lot of prior knowledge (in the algorithm design, parameter choices, etc.) that is not accounted for in the theory; finite state and actions is not a good model for many problems of interest; and a focus on unimportant quantities that nobody cares about.
    \item \textbf{Uneven Playing Grounds.} 
    Measuring the performance of an algorithm on a benchmark is confounded by the resources available to the implementer, such as proficiency in hyperparameter tuning, the size of the neural network trained, or prior knowledge about the problem/solution. The variability in the scale of experiments and software engineering affordable by different researchers (e.g., academia vs.~industry), and the current trend at top conferences to  prefer massive experimentation over conceptual novelty, can inhibit long-term progress.
    \item \textbf{Lack of Experimental Rigor.} Impressive singular experiments sometimes give a false sense of progress. While good PR click-bait, these may only disappoint as they preclude research on ``solved" problems which are in fact far from solved. For example, the simple 2D ProcGen Maze benchmark remains unsolved. We need more rigorous evaluation of difficulty and success. Moreover, for industry to adopt an approach, an impressive result is not enough -- stability, development time and cost, testability and life-cycle issues are critical. Currently, the publication standard is that failure cases are almost never reported, stability is impossible to tell, and software design issues are not even discussed.
\end{enumerate}

\section{Generalist Agents vs.~Deployable RL Systems}

There are two main dogmas in the RL community on how to make progress in solving real world decision making problems. The first, which we  refer to here as the `generalist agent' view (sometimes called ``RL first''), is that future progress will be made by focusing attention on large-scale training of agents that solve diverse problems, with the hope that along the way a generalist agent will develop, and will be a useful component in various real world problems. The second view, which we refer to as `deployable RL', takes a more pragmatic view seeking to design RL algorithms that solve concrete real world problems (sometime this view is called ``RL second''). 
\textit{The five problems described above are relevant to both approaches}. In the fix we propose, however, we focus on the second view. The reason is that with the current knowledge in the field, we believe there are concrete benefits to reap in deployable RL, while the generalist agent prospect is still at an early development stage.

\section{Principles of Deployable RL Research}

It is important to understand that at present, RL is uneconomical to deploy. Changing this requires both research of how to deploy RL effectively, and also a better understanding of the gains that an RL solution may bring, which will eventually make it worthwhile to pursue. We propose a constructive model for research in RL which we feel is relevant for our current state of knowledge in the field, and will advance progress towards deploying RL-based solutions in the real world. 
In the following, we outline three general principles. Our hope is that researchers will adopt some of these principles in their projects, and in turn, publication standards will grow to value these principles appropriately. 

\begin{enumerate}
    \item {\bf Challenges instead of benchmarks} A \textit{challenge} is a problem (real or not) sponsored by a group of researchers from academia/industry. Different from a benchmark, a challenge is not aimed at comparing algorithms, but is meant to simply be solved. That is, \textit{there is real world value in making progress on a challenge that holds regardless of the RL} (or other) algorithm used to solve it.

    {\bf Credit for contributing challenges} A rigorous presentation of a challenge should be credited as an important contribution (in terms of citations, impact, etc.). The contribution is not just the specific description of an application, but rather a community around it, and a supporting platform (code, scoreboard, etc.). We envision a special type of papers: ``contributed challenges". A fundamental criterion for evaluating the challenge would be a quantifiable measure of making \textit{real, and well-accepted progress} on it. That is -- no more maximizing made up rewards, but working towards indicators of real-world progress. 
    
    {\bf Measurable progress is the main criterion for publication}. Every publication that makes progress on a challenge either by suggesting a new algorithm, a positive or negative result, should explain the limitations and issues with the proposed new algorithm and how it addresses progress specifically. The main criterion for publishing a work is the progress made in a {\em quantifiable} manner.

    {\bf Weight class.} Compute available during development distinguishes contributors to ``weight class" which should also be reported. The overall amount of compute used for research is important to assess the significance of the results.  
    
    \item {\bf Theory papers should address specific challenges.} When a theory paper proposes a model, an algorithm, or a solution, the problem should be clear and grounded in a real challenge. We are not opposed to working on theory of small finite models but the goal of the research should be well justified in terms of its potential impact on real world problems. Theory papers should also consider problems that have to do with the system life cycle focusing on issues such as data acquisition, debugging, testability and performance deterioration. 
    
    \item {\bf Design-Patterns Oriented Research.} Software design is based on plugging multiple design patterns, massaging them, and repackaging a software system. A design pattern in software engineering is a skeleton of a solution for a particular problem that can be potentially reused for many instances. Similarly, real-world RL based systems should have conceptual solutions to problems where issues such as testability, debuggability, and other system life-cycle issues are addressed. Approaches to life-cycle management that solve problems grounded in applications are a welcome addition to RL research. Importantly, we foresee that one way to make significant progress on a challenge would be by developing novel design patterns for it.
\end{enumerate}

\section{Actionable Next Steps}

The sooner the community starts focusing on deployable RL, the better the chances  of seeing large scale real world impact. To us, deployable RL means RL that can work at scale, be economically feasible, and can eventually be put in the field.
If you relate to the points above, here's what you can do.

\begin{enumerate}
\item {\bf Contribute challenges}. Contributing challenges require deep understanding of an application domain. Some industrial research labs are already well positioned to start tackling this task. There are many scientific and engineering disciplines where RL could make an impact, such as the natural sciences, medicine, and manufacturing. A challenge should matter to the discipline it comes from so while the RL expert may not possess the domain knowledge, their goal in challenges is to explain to the domain experts what is possibly feasible and what is not. Joining forces with industry or other academia would lead to better framed and more realistic challenges.
\item {\bf Frame your own research.} A change of focus may be needed when writing papers and conducting research to focus on solving real problems that matter, and framing the research effort within the deployable RL principles. 
\item {\bf Criticize others' research.}
Affecting a change to the RL research community would likely require a coordinated effort from researchers, reviewers, and senior area chairs.
 If you are reviewing RL papers, shift gears from the common benchmark-driven evaluation, and ask how a paper gets the field closer to real-world impact. 
 If you are a senior reviewer or an area chair -- consider instructing your reviewers to judge papers differently.
\end{enumerate}

The online version of this manuscript (\url{https://avivtamar.substack.com/p/deployablerl}) contains a comment section. As a starting point, we invite researchers to propose and discuss ideas for challenges there.


\begin{thebibliography}{9}
\bibitem{lewin}
{Lewin, K. 1952. Field theory in social science. New York: Harper \& Row.}

\bibitem{mnih}
{Mnih, V., et al. 2015. Human-level control through deep reinforcement learning. Nature 518(7540), 529-533}
\end{thebibliography}
\end{document}